# Temporal Concept Drift and Alignment: An empirical approach to comparing Knowledge Organization Systems over time

Sam Grabus[1][0000-0003-4670-5690], Peter Melville Logan[2][0000-0003-2362-8282],

and Jane Greenberg[1][0000-0001-7819-5360]

[1] Metadata Research Center, Drexel University, Philadelphia PA 19104, USA
[2] Temple University, Philadelphia PA 19122, USA

## 1 Introduction

Historical humanities resources are increasingly available in digital form for scholarly study, with topical access provided through controlled vocabularies, such as the *Library of Congress Subject Headings* (LCSH). One key challenge in this area is that the language of a contemporary controlled vocabulary applied can differ drastically from the language used in the historical documents. As a result, key topics in the historical text may not be represented in the final metadata record. This happens because the historical terms are no longer part of our contemporary vernacular language, and, therefore not part of current knowledge organization systems (KOSs). The anachronistic relationship between a historical document and controlled vocabularies used is even more problematic when applying an automatic indexing approach for metadata creation. (Grabus et al. 2019; Logan et al. 2019). Research in this area is needed for two key reasons: first, to aid historians in finding materials in digital libraries and archives, and secondly, to allow information professionals better assist users by comprehensively contextualizing the space and time in which historical resources were created, through subject metadata. Research on how historical vocabulary versions can be used to extract temporally-aligned terms has not been extensively pursued. The work presented in this paper addresses this need, and contributes to the developing body of scholarly research about topical representation for historical humanities resources.

This paper reports on research comparing the automatic indexing output for a single digitized historical resource, using temporally-aligned and contemporary versions of a vocabulary to identify temporal concept drift. Background research was conducted as part of the "Developing the Data Set of Nineteenth-Century Knowledge" project[1], a National Endowment of the Humanities collaborative project between Temple University's Loretta C. Duckworth Scholars Studio and Drexel University's Metadata Research Center. The next section will discuss how the application of historical vocabularies can be used to enrich and contextualize humanities resources through temporally-aligned subject representation, as well as the exploring the zeitgeist and infrastructural inversion—ideas that examine the ways in which time period changes concepts. Another key research area reviewed includes automatic indexing with controlled vocabularies for humanities resources. The background section is followed by research objectives and methodology supporting the analysis. Next, the results are presented, followed by a contextual discussion, methodological contribution, and a conclusion that discusses key findings and future research.

## 2 Background

### 2.1 Temporal Concept Drift and Alignment

Concepts and vernacular language in humanities resources change over time. Parallel to this phenomenon, concepts in KOSs, such as controlled vocabularies, also change over time (e.g., across different editions of the Library of Congress Subject Headings). Several knowledge organization researchers have examined changes in KOSs, over time. Tennis (2002) describes the study of an indexing language concept, across vocabulary versions during its lifespan, as subject ontogeny. Tennis (2012) and Lee (2016) have both examined temporal changes and subsequent application of the concept “eugenics,” across the *Dewey Decimal System* (DDC), the *New Classification Scheme for Chinese Libraries*, and the *Nippon Decimal Classification*. Fox (2016) analyzed the subject ontogeny of intersex people, as the concept has been classified and represented across 23 editions of the DDC, which reflects the uncertainty and changing understanding of the concept apparent in the literature of the times.

Other researchers have explored the use of historical controlled vocabularies for representing terminology differences in historical resources. The PeriodO project examined pre-twentieth-century controlled vocabulary concepts that described time periods relevant to cultural heritage artifacts (Rabinowitz 2014). This project underscored the disagreements across vocabularies that were developed in a particular time and place. For example, “Iron Age” can refer to different dates and locations, depending on where and when that term was created in any given controlled vocabulary (para. 4). Dobreski et al. (2019) have also explored use of multiple period-specific terms to represent topics for a collection of nineteenth-century sideshow performer photographs. These researchers call for the use of multiple historical and contemporary vocabularies to contextualize concepts as they were represented at the time, as well as how they are represented in the twenty-first century. For example, their research demonstrates the representation of the concepts “Chinese midget” and “bearded lady” in the metadata records, and makes the case for also providing contemporary terms in the metadata records for the photographs to contextualize the changes in our knowledge organization systems (18).

The increased availability of controlled vocabularies in machine-processable format, such as linked data, increases opportunities to leverage these tools for automatic indexing. The challenge is that the vocabularies accessible in linked data are almost exclusively contemporary. Therefore, highly-relevant candidate terms may not be able to match to controlled vocabulary concepts for historical documents. In other words, if a nineteenth-century digital resource contains a term that no longer reflects contemporary literary warrant in the twenty-first century, that candidate term will have no chance of appearing in automatic subject indexing results, because it has been removed from the pool of authorized KOS terms.

Toepfer and Siefert (2018) suggest that the effect of this externally-caused temporal drift (referred to from here forward as temporal concept drift) could be addressed through the use of temporally-aligned controlled vocabulary version releases. In other words, if a temporally-aligned vocabulary that does contain the obsolete candidate term is applied to index the nineteenth-century resource, the term now has a greater than zero percent chance of appearing in the automatic subject indexing results. Toepfer and Seifert explored temporal concept drift over decades, with a dataset comprised of document titles and author-provided keywords. Their methodology provides a foundation for further research examining terminological drift across longer time periods and with more robust full-text datasets. Foundational work here illustrates the notion of KOSs capturing the zeitgeist, and supports the solution of infrastructure inversion, as discussed below.

## 2.2 KOS Zeitgeist and Infrastructural Inversion

In 1968, Patrick Wilson described bibliographic control as a form of power (6). Similarly, Bowker and Starr's (1999) describe classification systems as a reflection of the dominant worldview within a particular space and time, or zeitgeist. As KOSs capture the zeitgeist, KO researchers such as Sanford Berman (1971, 19) have examined the ways in which "inherited assumptions and underlying values" have manifested in our knowledge organization systems. Olson (2002, 15) describes the classifying or naming of information objects as a structuring of reality that is inherently biased and disenfranchising, based upon the inequities inherent through literary warrant.

Adler et al. (2017) propose studying specific concepts in our knowledge organization systems as a means to turn society's gaze back upon itself, in recognition that it is impossible to completely erase bias from our KOSs. An example is found in Bowker and Starr's (1999) approach of infrastructural inversion, which is a means of examining systemic erasure of social and political worldviews within classification systems. They describe infrastructural inversion as "a struggle against the tendency of infrastructure to disappear" (34). They argue that infrastructural inversions can identify the constraints of historically-contingent infrastructure, as well as its ramifications (34).

Infrastructural inversion has primarily been pursued through subject ontogeny case studies for individual concepts. There is a need for large-scale examination of infrastructural inversion to more fully highlight systemic erasure of social and political views. Contemporary KOSs that go as far back as the 19th century, such as the Library of Congress Subject Headings, present an opportunity for large-scale examination, which is the overriding goal of this paper. This research goal, and the overarching need to scale up the use of controlled vocabularies for indexing large digital humanities collections, necessitates the use of automated approaches.

## 2.3 Automatic Indexing for the Humanities

Automatic indexing broadly describes the process of applying computational approaches for representing resources, often with controlled terms. Automatic indexing has not progressed as quickly in the humanities and social sciences domains as compared to scientific domains, largely due to challenges of less funding. Additional challenges include the linguistic indeterminacy that is inherent in humanities resources, low levels of specificity and precision, and the importance of rich historical context for representing these resources (Agosti et al. 2014; El-Haj et al. 2013; Svenonius 2000). Humanities and social sciences resources also span across multiple domains of knowledge. Some multidisciplinary corpora, such as historical encyclopedia entries, require the use of a controlled vocabulary that extensively covers every domain of knowledge. As a result, digital libraries may forgo the use of controlled vocabularies altogether. Golub et al. (2020) evaluated the application of indexing terms for humanities resources across Scopus and a local public university repository in Sweden. Their bibliographic analysis discovered that only 13.1% of the humanities articles were indexed using a controlled vocabulary, none of which were humanities-specific or even multidisciplinary, and a need for highly-granular indexing terms for humanities resources was identified (1206). The research reviewed above points to the need for further research on automatic indexing in the humanities, and informs the goals and questions for the research presented in this paper.

## 3 Goals and Research Questions

This research presents a methodology for examining how the temporal alignment of controlled vocabularies can be used to highlight temporal concept drift in our KOSs. To study this, two controlled vocabularies were selected: the 1910 version of LCSH and contemporary FAST Topical headings. The comparison focused on indexing output for each vocabulary for the selected nineteenth-century corpus.

Specific research questions that were pursued are:

1. In examining the top terms retrieved with 1910 LCSH and the top terms retrieved with FAST Topical from an automatic indexing sequence, which results are exclusive to the 1910 LCSH output?
2. Taking the exclusive set from the 1910 LCSH results from question one, which terms are deprecated from the authorized terms in the contemporary LCSH vocabulary version (2020 FAST Topical)?
3. Given a selected subset of these deprecated terms, what is the contemporary counterpart term that can be identified in the current FAST Topical vocabulary?

A research sample of nineteenth-century *Encyclopedia Britannica* entries, made available through the Nineteenth-Century Knowledge Project[2], provides a use case for answering these research questions. The research questions were investigated through comparison and basic descriptive statistics. Tools used to answer these research questions consist of HIVE (Helping Interdisciplinary Vocabulary Engineering)[3], 1910 Library of Congress Subject Headings, and 2020 FAST Topical, which will be discussed in more detail in the next section. The next section of this paper will report on the methods and steps employed to answer the research questions.

## 4 Methods and Procedures

This research was approached through a comparison of subject headings automatically generated with two versions of a controlled vocabulary: a temporally-aligned nineteenth-century version (1910 LCSH), and a contemporary 2020 version (2020 FAST Topical). The protocol for performing this study involved the following steps:

1. Compile a sample for testing:
   1.1. A stratified random sample of 90 encyclopedia entries was selected from a convenience sample of 19,912 *Encyclopedia Britannica* entries across four historical editions of the *Encyclopedia Britannica* the 3rd, 7th, 9th, and 11th editions (spanning the years from 1788-1911). These four editions combined consist of over 100,000 entries.
   1.2. Entry length: In order to account for the wide range of encyclopedia entry lengths in the convenience sample, three sets of samples were used to represent short, medium, and long entries. This decision was made based upon previous findings (Grabus 2020) that encyclopedia entry length affects the quantity of automatic indexing results, reflecting the availability of candidate terms in the entry. The following parameters were used:
      - Short: 100-2,000 words

- Medium: 2,001-99,000 words
- Long: Greater than or equal to 100,000 words

The short and medium entry samples were comprised of forty entries each, while the long entry sample contained ten entries, since only ten entries in the convenience sample fit the entry length criteria.

2. Perform automatic subject indexing sequence:

2.1. Full text and NER (named entity recognition) entity documents:

2.1.1. Each of the ninety entries in the sample were represented through two separate documents, for a total of 180 documents: full text entries and extracted entity types. These extracted entities were used as a complementary experimental approach to automatically index this corpus, as part of ongoing research.

2.1.2. Named entities were extracted using Stanza[4], Stanford NLP's current natural language processing toolkit for Python, which includes Stanford NER[6]. Stanza was used to generate and extract the following entity types: Nationality, religions, political organizations; people; dates; languages; artworks; events; laws; products; organizations; and facilities.

2.2. 2020 FAST Topical and 1910 LCSH controlled vocabulary versions.

2.2.1. The topical facet of the FAST (2020 FAST Topical) vocabulary was selected as the contemporary version of LCSH for this study. FAST (Faceted Application of Subject Terminology) is a simplified and faceted version of the Library of Congress Subject Headings, developed by OCLC[5]. The multi-disciplinarity and exhaustivity of LCSH have contributed to its continued use across digital library collections (Walsh 2011), and FAST helps to facilitate automated approaches because of the simpler structure and syntax, particularly through the reduction of precoordinated subject strings in favor of single-concept terms (O'Neill and Chan 2016). FAST was selected for this research to address common challenges of applying controlled vocabularies through automatic indexing, which include redundancy of terms, low levels of specificity, free-floating subdivisions, and the nested pre-coordination of terms (Golub et al. 2020; O'Neill and Chan 2016; Khoo et al. 2015; El-Haj et al. 2013).

2.2.2. The first printed edition of LCSH (LCSH 1910) was selected as the historical vocabulary counterpart for this study. Development of this first edition began in 1909 as "Subject Headings Used in the Dictionary Catalogues of the Library of Congress" (Stone 2000). It was published in parts, between the years 1910 and 1914[6]. While heavily influenced by the subject access guidelines established by Charles Cutter, Library of Congress cataloguer J.C.M Hanson adopted an approach that was more practical than philosophical, which led to early inconsistencies in LCSH cross-references and subheadings (Foskett 1996). To date, the historical LCSH version has not been developed for automatic approaches. The Metadata Research Center has led an effort to prepare the vocabulary by converting it into a machine-readable format for automated use.

2.3. Fixed recall: A fixed maximum recall for indexing results was set to ten per each vocabulary: ten maximum terms returned with 1910 LCSH, and ten terms returned with 2020 FAST Topical.

2.4. Automatic indexing tool: The automatic subject indexing for this study was performed with HIVE (Helping Interdisciplinary Vocabulary Engineering), a vocabulary server and automatic indexing application (Greenberg et al. 2011). HIVE allows the user to upload a digital text or website URL, select one or more controlled vocabularies, and perform an automatic subject indexing sequence to extract natural language keywords and match them to controlled vocabulary terms through stemming and wildcards. The HIVE tool integrates the RAKE (Rapid Automatic Keyword Extraction) algorithm (Rose et al. 2010).

3. Data collection: An outer merge was performed on indexing results to isolate which results only appear in the 1910 LCSH results. These 1910 LCSH results were then manually searched in the full subject heading list for 2020 FAST Topical to determine which terms no longer exist as an exact match in the contemporary vocabulary counterpart.

4. Analysis:

4.1. Basic descriptive statistics were generated to determine the percentage of terms that only appear in the 1910 LCSH indexing results, and what percentage of the 1910 results no longer appear in the 2020 FAST Topical vocabulary version, demonstrating temporal drift.

4.2. Results were compared across indexing approaches and encyclopedia entry length.

4.3. Authorized and variant terms were also identified among the results.

4.4. For a selected subset of the concepts representing temporal drift, the FAST 2020 counterpart concepts that have likely replaced them were manually identified through mapping the 1910 LCSH concepts to variant terms in 2020 FAST. The terms selected for this portion of the analysis were chosen to reflect common themes that have emerged from preliminary observation of the results.

## 5 Results

The results report three aspects of using the 1910 LCSH to generate subject headings for the historical encyclopedia entries: 1) the portion of results that are exclusive to the 1910 LCSH output; 2) from the exclusive set of 1910 LCSH results from question one, the portion of terms that are deprecated from authorized use in the contemporary vocabulary version, demonstrating temporal concept drift; and 3) given a selected subset of these deprecated terms, what is the contemporary counterpart term that can be identified in the current FAST Topical vocabulary.

Table 1 provides a high-level view of results for these first two research questions. Across the 1,478 total automatic subject indexing results for 180 total sample files, 31 percent of the 1910 LCSH results do not appear in the 2020 FAST Topical results, and 7.24 percent of the 1910 LCSH results represent terms that no longer exist in the contemporary vocabulary version as an exact match, demonstrating temporal concept drift.

**Table 1**: Percentage of terms exclusive to the 1910 LCSH results, and percentage of 1910 LCSH terms that are deprecated from the authorized terms in the contemporary FAST Topical, demonstrating temporal drift.

| | |
|---|---|
| **Total Documents** | 180 |
| **Total 1910 LCSH Indexing Terms** | 1478 |
| **Terms Exclusive to 1910 LCSH Output** | 458/1478 (31%) |
| **Terms Demonstrating Temporal Drift** | 107/1478 (7.24%) |

These results were also compared by indexing approach, encyclopedia entry length, and the comparison of authorized and variant terms in the results. Table 2 compares results by the two indexing approaches: indexing with the full text of the encyclopedia entries, and indexing concepts that were extracted using the NER-to-ontology mapping approach. When examining the portion of results representing temporal drift across each approach, the full text approach demonstrated 6.2 percent, and the NER-to-ontology approach demonstrated 8.78 percent.

**Table 2**: Percentage of exclusive 1910 LCSH results and temporal concept drift as it differs across the two indexing approaches applied for this research: full text and NER-to-ontology mapping.

| **Indexing Approach** | **Full Text** | **NER** | **Both** |
|---|---|---|---|
| **Number of Documents** | 90 | 90 | 180 |
| **Total 1910 LCSH Indexing Terms** | 886 | 592 | 1478 |
| **Terms Exclusive to 1910 LCSH Output** | 277/886 (31.26%) | 181/592 (30.57%) | 458/1478 (31%) |
| **Terms Demonstrating Temporal Drift** | 55/886 (6.20%) | 52/592 (8.78%) | 107/1478 (7.24%) |

**Table 3**: Percentage of exclusive 1910 LCSH results and temporal concept drift as it differs across the three samples representing entries of different lengths.

| **Entry Length** | **Short** | **Medium** | **Long** | **All** |
|---|---|---|---|---|
| **Total 1910 LCSH Indexing Terms** | 542 | 736 | 200 | 1478 |
| **Terms Exclusive to 1910 LCSH Output** | 171 | 223 | 64 | 458 |
| **Terms Demonstrating Temporal Drift** | 34/542 (6.27%) | 57/736 (7.74%) | 16/200 (8%) | 107/1478 (7.24%) |

Table 3 displays the results by sample, each of which included entries of different lengths. The proportion of terms representing temporal drift for each sample were as follows: short entries, 6.27 percent; medium-length entries, 7.74 percent; and the longest entries, 8 percent, demonstrating a slight increase as entry length increases. In order to understand the nuances and complexities of the 1910 LCSH dataset, Table 4 separates the portion of terms representing temporal drift by authorized term and variant terms. Of the 107 total results representing temporal drift, 69.16 percent were headings listed as authorized terms in the 1910 LCSH, while 30.84 were variant terms that cross-referenced to an authorized term. These variant terms were included in the results if the variant term or its authorized form no longer exist as an exact match in the 2020 FAST/LCSH. The full table of results is displayed below in Table 5.

**Table 4**: Distribution of authorized and variant headings across the 1910 LCSH terms representing temporal concept drift.

| | |
|---|---|
| **Authorized Term Results** | 74/107 (69.16%) |

| **Variant Term Results** | 33/107 (30.84%) |
|---|---|

**Table 5**: Full table of results, compared by encyclopedia entry length, indexing approach (full text vs. NER)

| **Sample** | 1 | 2 | 3 | 1 | 2 | 3 | **Total Across Samples** |
|---|---|---|---|---|---|---|---|
| **Encyclopedia Entry length** | Short | Medium | Long | Short | Medium | Long | N/A |
| **Indexing Approach** | Full Text | Full Text | Full Text | NER | NER | NER | N/A |
| **Number of Documents** | 40 | 40 | 10 | 40 | 40 | 10 | 180 |
| **Total 1910 LCSH Indexing Terms** | 400 | 386 | 100 | 142 | 350 | 100 | 1478 |
| **Terms Exclusive to 1910 LCSH Results** | 137/400 (34.25 %) | 113/386 (29.27%) | 27/100 (27%) | 34/142 (23.94%) | 110/350 (31.43%) | 37/100 (37%) | 458/1478 **(30.98%)** |
| **Authorized Terms Demonstrating Temporal Drift** | 16 | 16 | 3 | 5 | 26 | 8 | 74 |
| **Variant Terms Demonstrating Temporal Drift** | 7 | 9 | 4 | 6 | 6 | 1 | 33 |
| **Total Terms Demonstrating Temporal Drift** | 23/400 (5.75%) | 25/386 (6.48%) | 7/100 (7%) | 11/142 (7.75%) | 32/350 (9.14%) | 9/100 (9%) | 107/1478 **(7.24%)** |

Seven preliminary results were identified as not existing in 2020 FAST Topical, but existing in 2020 LCSH, as a result of using just the faceted topical subset of LCSH. These seven results were also removed from the final results, as they do not represent temporal drift as it has been defined for this study. Additionally, a total of five terms that were erroneously included in the initial pool of results were identified as inaccuracies related to optical character recognition (OCR) and regular expression inconsistencies inherent throughout the 1910 LCSH cross-references and subheadings (Foskett 1996); they were subsequently removed from the final results.

Table 6 addresses research question three by isolating a subset of terms that demonstrate temporal drift, and identifying the contemporary counterpart concept that has likely replaced it as the authorized heading. The terms selected for this table were manually chosen from the 107 terms demonstrating temporal concept drift. Terms in the left column represent 1910 LCSH subject headings that no longer exist as an authorized term exact match in the contemporary vocabulary version, having fallen out of use since 1910 LCSH. Terms in the right column are the current 2020 FAST Topical subject headings that have replaced the nineteenth century terms as the authorized heading, as indicated through the listed variant terms. The terms selected for this table were manually chosen to reflect common themes that have emerged from preliminary observation, which include religion, nationality/ethnicity, science, and gender.

**Table 6**: An excerpt of 1910 LCSH concepts representing temporal drift, alongside the verified or probable replacement terms in 2020 FAST Topical/LCSH.

| 1910 LCSH Subject Heading | 2020 FAST Topical Subject Heading |
|---|---|
| Mohammedans | Muslims |
| Saracens | Islamic Empire |
| Moors (The race) | Muslims |
| Gipsies | Romanies |
| Uzbegs | Uzbeks* |
| Scotch | Scots |
| Omayyads | Umayyad dynasty |
| Malay Race | Malays (Asian people) |
| Gallas | Oromo (African people) |
| Polyzoa | Bryozoa |
| Man | Human beings |

*1910 LCSH term is not listed among variant terms for the likely 2020 FAST counterpart.

Overall, the results in Tables 1-4 show that a significant number of LCSH terms were only retrieved during the automatic indexing process when using the 1910 LCSH vocabulary. Some of these terms exclusive to the 1910 LCSH indexing output have since been replaced by the Library of Congress with a different term. Table 5 provides examples of these replacements to illustrate this temporal concept drift. These results help point to important issues regarding topical representation for historical resources, which are discussed below.

## 6 Discussion

The research above presents a comparison of subject headings generated for the nineteenth century encyclopedia entry sample using historical and contemporary versions of the Library of Congress Subject Headings (1910 LCSH and 2020 FAST Topical). While researchers have discussed the potential for using temporally-aligned vocabularies to contextualize historical resources, there is limited work on applying these historical terminologies to a corpus automatically, and comparing the results to highlight temporal concept drift and contextualize the historical resources.

The research demonstrated that for a sample of 90 encyclopedia entries from the research sample, 31 percent of the total 1910 LCSH indexing output terms were exclusive to the 1910 LCSH output. Further analysis found that 7.24 percent of all 1910 LCSH indexing results comprised terms that are no longer found in the current vocabulary counterpart, demonstrating temporal concept drift. The proportion of temporal concept drift was slightly higher in results generated using the indexing approach that employed NER-to-ontology mapping. The proportion also slightly increased using samples with longer encyclopedia entries, which is likely an effect of increased overall indexing results for the sample entries longer than 2,000 words.

When comparing the full text and intermediary NER indexing approaches, the output demonstrated that each approach provides almost entirely different sets of output terms. While the comparison between NER and full text indexing approaches is not the central focus of this research, this finding will be explored further in ongoing dissertation research.

The results also highlighted examples of the ways that concept drift manifests in the indexing results for the encyclopedia entries, and the subject headings that have replaced them over the course of a century. These 1910 LCSH terms represent headings that only exist in the 1910 LCSH vocabulary version, and have since fallen out of use, or no longer appear as an exact match in the 2020 FAST Topical vocabulary version. The types of changes range from slight grammatical shifts to entirely new terms. For example, the 1910 LCSH terms “man” and “woman” have changed over time, with the authorized headings of “human beings” and “women” are now used in 2020 FAST Topical instead (note: “man” is not officially listed as a variant term under 2020 FAST Topical authority file for “men”). Demonstrating a more significant shift, the 1910 LCSH term “Gipsies” has been removed entirely, as it is considered a racial slur against the Roma people. Similarly, the subject heading “Moors (the race),” which is largely viewed as an archaic and pejorative term in 2022, has been replaced with the authorized subject heading “Muslims.” These examples reflect the ways in which many identity-based concepts of nationality/ethnicity and religion have changed over the last century, shining a spotlight upon the zeitgeist as it manifests through both the Library of Congress Subject Heading/FAST versions and the indexed historical encyclopedia entries.

The reported results are specific to this particular corpus of nineteenth-century encyclopedia entries, but it is very likely that this approach could be applied to other historical documents, across many domains of knowledge. For example, possible future applications could include existing nineteenth century digital library collections that would benefit from the addition of topical metadata, such as the Mark Twain Project letters[7]. This study exclusively examined the top ten indexing results, but future research can expand to evaluate the top twenty results, with the addition of relevance evaluation to ensure validity, and substantive qualitative analysis to highlight trends across the results. Next steps also include refining the conversion of the 1910 LCSH to reduce the presence of errors related to processing the vocabulary for computational access.

## 7 Methodological Contribution

In moving forward, there is limited research that outlines how to study temporal concept drift in KOSs at a larger scale. This paper presents a methodology that is guiding a larger study and provides a way that other researchers may compare terminologies or other historical vocabulary versions. Specifically, this methodology can be used to study at a larger scale how KOSs reflect social and political views at a particular point in time, and how the examination of changes and conceptual erasures in KOSs can be used to shed light upon inherent power structures and biases within society.

With any methodology, there are limitations, particularly due to the diversity of digital humanities collections, subject-appropriate terminologies, and tools for pursuing this approach. There is no one-size-fits-all solution for automatic or semi-automatic indexing, particularly within the humanities and social sciences. The methodology used in this research can be used to identify project-specific obstacles that need to be addressed, or adaptations that can be made to examine temporal concept drift for domain-specific vocabularies and historical collections.

## 8 Conclusion and Next Steps

The objective of this research was to examine the use of temporally-aligned and contemporary vocabulary versions to analyze temporal concept drift in both knowledge organization systems and digital humanities resources. Specifically, this study determined the prevalence of temporal concept drift apparent when using an historical vocabulary to extract subject headings representing controversial, obsolete, or otherwise changed terms in our KOSs, as it manifests through automatic application to the nineteenth-century *Encyclopedia Britannica* research corpus.

The significance of this research is that it demonstrates that historical controlled vocabulary releases can be used to generate anachronistic subject headings that represent temporal concept drift in both KOSs and historical resources. This approach can be applied as a complementary topical indexing approach to contextualize historical resources using the language of the time. This approach has the potential to turn society's gaze back upon itself, through the way that topical metadata is generated for historical digital library resources. This approach could also be harnessed to map these anachronistic subject headings to their contemporary counterpart subject headings, through variant term matching. This would allow information professionals to have the best of both worlds, through the extraction of temporally-aligned historical concepts and the final representation with its counterpart authorized contemporary terms.

The research presented here opens up a pathway for further historical research across multiple domains, ranging from the humanities to science. It also informs research in the digital humanities and information sciences, as they intersect with knowledge organization. As already stated, future research is underway, which includes validity checking through relevance evaluation, and the qualitative coding of concepts representing temporal drift, to identify whether particular conceptual themes arise as predominant. This research helps to address the overarching goal of understanding the challenge of KOS temporal concept drift and possibilities for KOS temporal alignment.

[1] https://securegrants.neh.gov/publicquery/main.aspx?f=1&gn=HAA-261228-18

[2] https://tu-plogan.github.io/

[3] https://hive2.cci.drexel.edu/

[4] https://stanfordnlp.github.io/stanza/

[5] https://www.oclc.org/research/areas/data-science/fast.html

[6] https://www.loc.gov/catdir/cpso/illegal-aliens-decision.pdf

[7] https://www.marktwainproject.org/